\documentclass{article}
\usepackage{spconf,amsmath,epsfig}
\usepackage{xcolor}

\newcommand{\pparagraph}[1]{\medskip\noindent\textbf{#1}}

\title{Portraying the Need for Temporal Data in Flood Detection\\ via Sentinel-1}

\name{~\hfill Xavier Bou$^1$ \hfill Thibaud Ehret$^1$ \hfill Rafael Grompone von Gioi$^1$ \hfill Jérémy Anger$^{1,2}$ \hfill ~}
\address{$^1$ Université Paris-Saclay, CNRS, ENS Paris-Saclay, Centre Borelli, France \hspace{2em} $^2$ Kayrros SAS}

\begin{document}
%
\maketitle
\begin{abstract}
Identifying flood affected areas in remote sensing data is a critical problem in earth observation to analyze flood impact and drive responses. While a number of methods have been proposed in the literature, there are two main limitations in available flood detection datasets: (1) a lack of region variability is commonly observed and/or (2) they require to distinguish permanent water bodies from flooded areas from a single image, which becomes an ill-posed setup. Consequently, we extend the globally diverse MMFlood dataset to multi-date by providing one year of Sentinel-1 observations around each flood event. To our surprise, we notice that the definition of flooded pixels in MMFlood is inconsistent when observing the entire image sequence. Hence, we re-frame the flood detection task as a temporal anomaly detection problem, where anomalous water bodies are segmented from a Sentinel-1 temporal sequence. From this definition, we provide a simple method inspired by the popular video change detector ViBe, results of which quantitatively align with the SAR image time series, providing a reasonable baseline for future works.
\end{abstract}

\section{Introduction}
\label{sec:intro}
\begin{figure}
    \centering
    \includegraphics[width=0.325\linewidth,trim={50 0 50 0},clip]{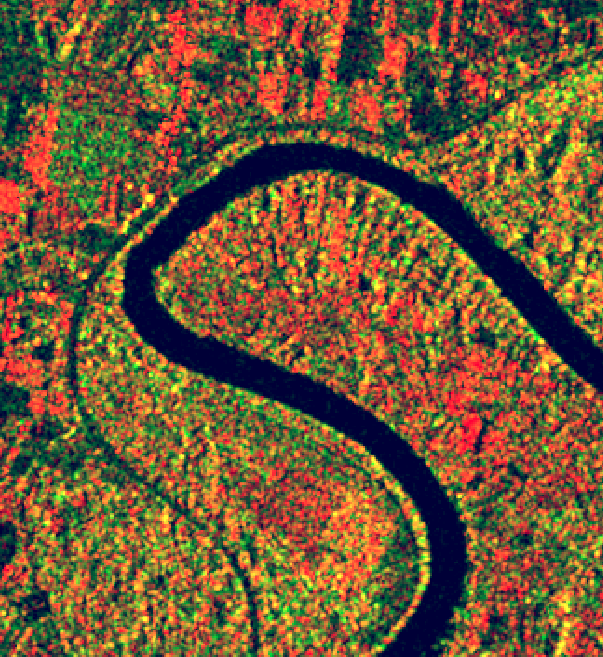}
    \includegraphics[width=0.325\linewidth,trim={50 0 50 0},clip]{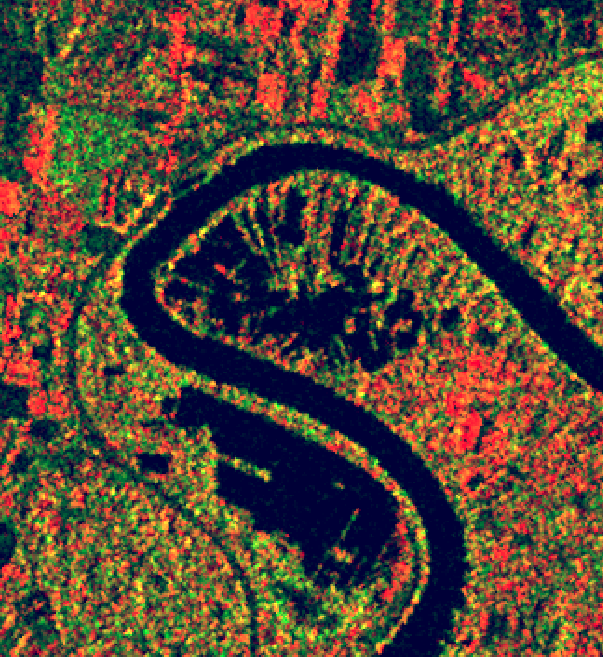}
    \includegraphics[width=0.325\linewidth,trim={50 0 50 0},clip]{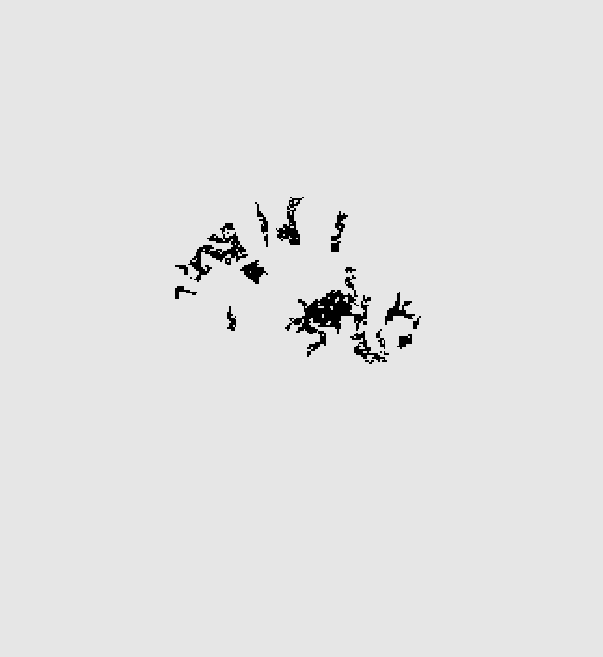}
    \caption{Sample from the MMFlood dataset~\cite{mmflood} (EMSR358-0-6). The left image was acquired on 2019/05/10 (not part of MMFlood), the middle image is from 2019/05/16 during the flood event. On the right, the MMFlood label shows only a partial annotation of the flooded areas. Note that from only the middle image it is not possible to infer which are the permanent bodies, a multi-date input is essential for flood mapping.}
    \label{fig:mmflood-limitations}
\end{figure}
\begin{figure*}[t]
    \centering
    \includegraphics[width=1\linewidth,trim={0.5cm 4.5cm 0.5cm 4.25cm},clip]{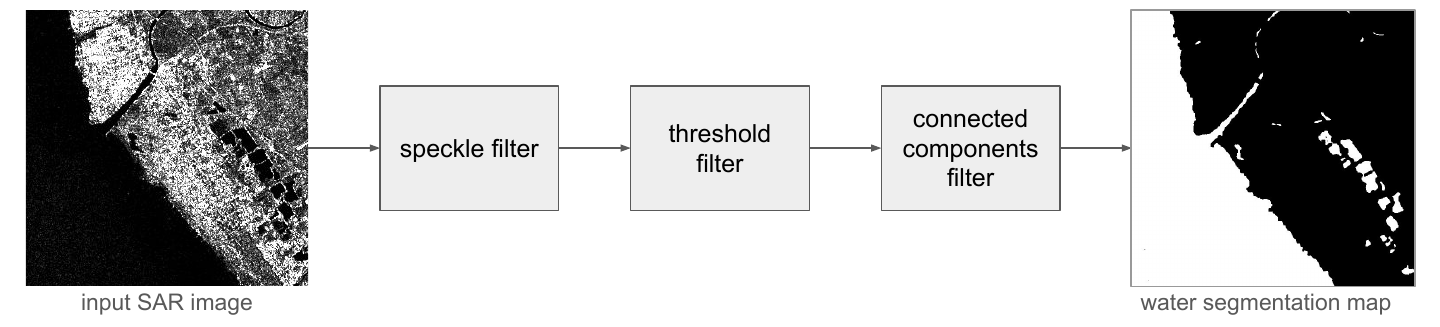}
    \caption{Water segmentation process. First, a speckle filter is used to denoise the raw SAR image. Then, a threshold filter is applied to generate a binary segmentation. Lastly, only connected components of relevant size are kept.  }
    \label{fig:water_segmentation_diagram}
\end{figure*}
According to the Centre for Research on the Epidemiology of Disasters (CRED), flooding constituted the greatest impact on global populations compared to any other natural disaster in 2020~\cite{cred20}. Moreover, the magnitude of future flood-related losses is expected to continue to increase in a number of regions around the globe~\cite{europe_predictions}. Consequently, there is a vested interest to leverage available earth observation resources to plan and accelerate the response to these events, and to analyze and predict their impact. Hence, automatic flood mapping, which attempts to segment water bodies corresponding to flood events and distinguish them from permanent water areas, is regarded as a crucial problem in the remote sensing community.

In recent years, the increasing availability of remote sensing data inspired the community to drive efforts to this issue. Due to its ability to penetrate through clouds, most flood detection methods leverage synthetic-aperture radar (SAR) data~\cite{review}, either via classical thresholding approaches~\cite{method_both_threshold_and_supervised,method_threshold_1} or modern deep learning approaches~\cite{dl_review, method_both_threshold_and_supervised}. Furthermore, some methods use additional optical information~\cite{method_multimodal_1, method_multimodal_2, method_multimodal_3}. Nevertheless, current available flood detection datasets bear at least one of the following two limitations. On one hand, a large number of proposed datasets contain a low number of events and thus lack region variability, which affects generalization on out-of-distribution data. On the other hand, a critical predicament is often embedded into flood detection datasets: temporal information is ignored and thus the detection is inferred from a single image. Distinguishing flooded areas from permanent water bodies from a single-image is an ill-posed problem. Moreover, ground-truth labels are often derived from existing flood monitoring services such as the Copernicus Emergency Management Service (CEMS), that offer mapping extracted from many sources during an event. As such, the labels are often inconsistent with the water extent observed from a given SAR time series, due to the different acquisition scenario (different acquisition date, orbits, resolution, etc.).

In this article, we analyze the most relevant flood detection datasets and highlight the mentioned limitations, illustrating the ill-posed nature of the problem. Furthermore, we select a recent and well distributed dataset, MMFlood~\cite{mmflood}, and extend it from single image to multi-date by providing image time series covering a year around the flood event. When comparing the image time series with the provided ground-truths, we identify that not all non-permanent water areas are annotated as flooded regions and some annotations are extraneous, resulting in an unclear definition of flooded pixel.

Consequently, we frame the task as a time series anomaly detection problem, where anomalous water areas should be segmented given previous observations. To this end, we provide an unsupervised method inspired by the popular video change detection algorithm ViBe~\cite{vibe}. A qualitative analysis indicates a more aligned performance with human interpretation when considering past observations. We believe our method can serve as a baseline for future analysis.

\section{The problem with flood detection datasets}\label{sec:datasets}
In order to train learning-based methods and evaluate models, recent flood detection datasets have been published.
Sen1Floods11~\cite{sen1floods11} and MMFlood~\cite{mmflood} are both single-date multimodal datasets.
MMFlood is constructed around 95 CEMS events and use the closest Sentinel-1 image to the event, a hydrography map and a DEM as inputs. The ground-truth labels are derived from CEMS reports, compiled by experts using various sources.
We argue that this dataset presents an ill-posed view of the flood detection problem. Indeed, as only one date is provided as input, methods -- traditional or learning-based -- cannot infer which are the permanent water bodies from the flooded areas. 
Furthermore, the dataset sometimes includes annotations of regions which seem unaffected when inspecting the Sentinel-1 image.
Figure~\ref{fig:mmflood-limitations} shows an example of partially annotated flooded areas. It is clear that from the single image in the middle, it is not possible to determine which water bodies are part of the flood extent; for a well-posed flood detection problem, it is necessary to use a multi-temporal approach.
This discrepancy between the observed imagery and annotated labels is likely due to the nature of the CEMS which provides rapid assessment by analyzing multiple satellite acquisitions.

Sen12-flood~\cite{sen12flood} offers a dataset comprised of Sentinel-1 and Sentinel-2 time series for each event. The labels are one boolean per date, indicating whether flooded areas are present in each image. These labels were derived from the CEMS. While the multi-date aspect of the dataset offers a better posed problem, the labels themselves only allow for flood detection on crops and are unsuitable for flood mapping.

\section{Extending a single image dataset}
The MMFlood dataset~\cite{mmflood} is well distributed and representative of a large set of flood events. To cope with the lack of temporal information, leading to the ill-posed definition of flood, we extend the MMFlood dataset by adding the Sentinel-1 images one year before the event and one month after.
To do so, we fetch the imagery from the same relative orbit number of the original selection of MMFlood, and process the Sentinel-1 GRD imagery to obtain the same footprint and radiometry. Due to the improved geolocation accuracy of our pipeline and different DEM, there are some negligible residuals between our dataset and the original MMFlood Sentinel-1 crop.
However, as discussed in Section~\ref{sec:datasets}, the labels of MMFlood are not suitable for a fair evaluation of flood detection methods, and we discourage their use.
The proposed multi-date dataset can be used for multi-date flood mapping methods but future work is necessary to include validation labels.

\section{Proposed algorithm}
\begin{figure*}[t]
    \centering
    \includegraphics[width=1\linewidth,trim={0.5cm 7.25cm 2cm 0.5cm},clip]{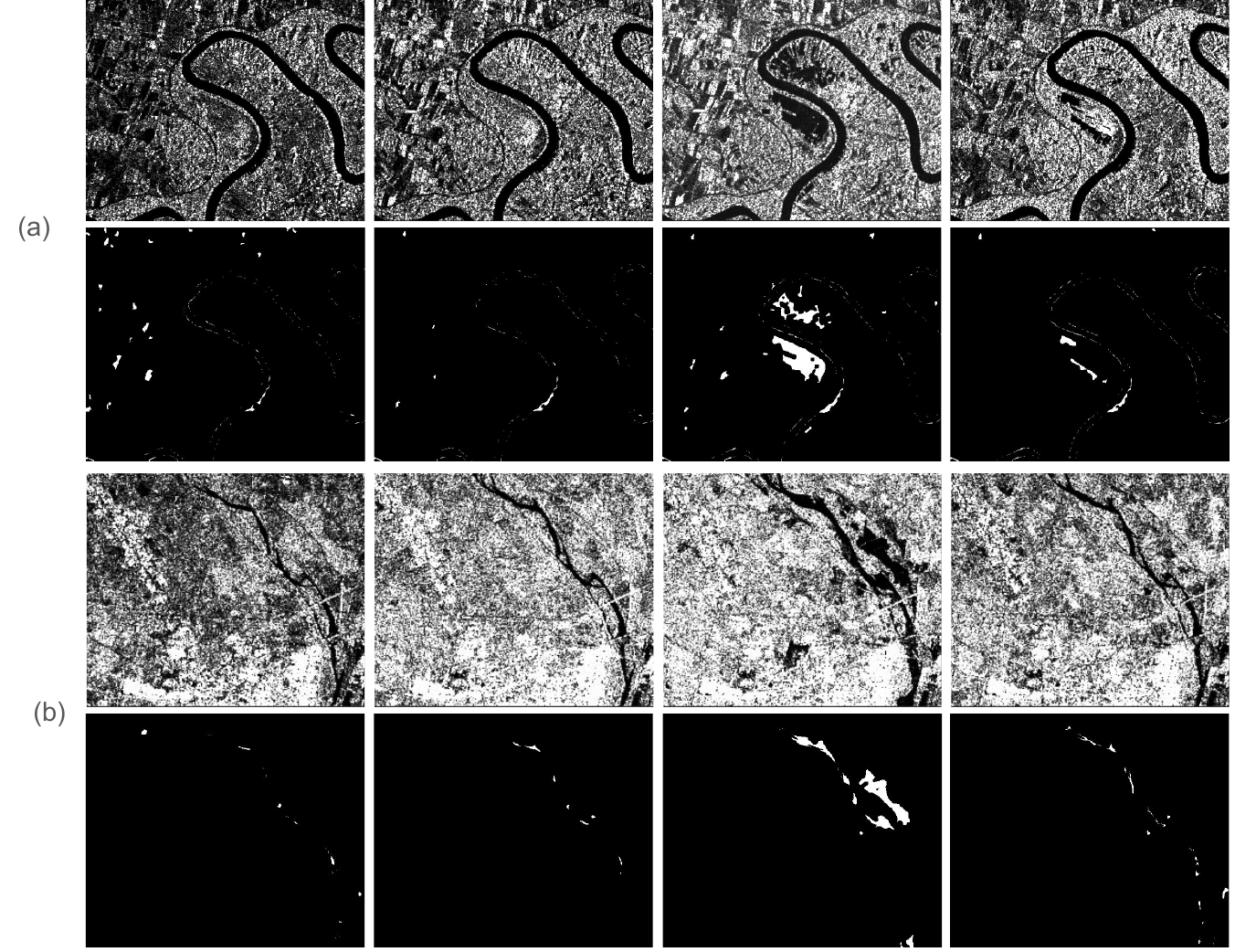}
    \caption{Qualitative results of the proposed unsupervised method for MMFlood scenes EMSR358-0-6 (a) and EMSR468-0-1 (b). Four observations during the flood event are shown for each, containing two observations prior to the event and two afterwards. The input SAR images are shown on the top row of the examples, while the output flood segmentation maps are displayed on the bottom.}
    \label{fig:qualitative_results}
\end{figure*}
Due to the inconsistent definition of flood witnessed in the evaluated datasets, we propose to re-frame the problem of flood detection from a temporal anomaly detection point of view. Also referred to as change detection, these types of methods build a statistical background model of the scene based on past observations~\cite{vibe, subsense}. Then, temporal anomalies are detected when a new observation differs from samples in the background model. Inspired by the popular change detection method ViBe~\cite{vibe}, widely used in the industry for its simplicity and real-time capabilities, we derive a flood detection approach to segment anomalous water regions in a Sentinel-1 time series. The two main steps of the proposed method are described in the following sections.
\subsection{Water segmentation}
Aiming to distinguish permanent water areas from anomalous flooded regions implicitly entails discriminating water pixels from land pixels. Hence, the first step of our pipeline is to generate water segmentation maps. For an input image $I\in{\rm I\!R}^{2\times H\times W}$ with height $H$, width $W$ and two polarization channels corresponding to VV and VH, respectively, we generate a binary segmentation map $I_b\in{\{0,1\}}^{H\times W}$ classifying pixels into \textit{water} or \textit{ground} clusters. This is done via a denoising and a thresholding process as illustrated in Figure~\ref{fig:water_segmentation_diagram}.

\pparagraph{Denoising. } Speckle noise, caused by random interference from scaterers in Earth surface~\cite{speckle_noise}, affects the segmentation process, producing both false negatives and false positives. A speckle filter is thus applied to the SAR image with the purpose of reducing image noise. It consists of a simple boxcar filter that smoothes the signal by applying a $(kernel\_size \times kernel\_size)$ sliding window.

\pparagraph{Thresholding. } A binary map $I_b$ is subsequently generated from the denoised SAR image $I_d$ as follows. First, a simple thresholding is applied so that
\begin{equation} \label{eq:threshold}
I_b(x) = \begin{cases}
            1,  & \text{if } I_d(x) > threshold \\
            0,  & \text{if } I_d(x) \leq threshold
\end{cases},
\end{equation}
where $x$ corresponds to pixel location. Secondly, the segmented image $I_b$ is further processed by removing small regions of inter-connected pixels. To this end, sets of connected components with less than $num\_components$ components are filtered out. We use a similar implementation from Grompone \textit{et al.}~\cite{ground_vis} to efficiently compute 4-connectivity regions in a binary image. The resulting segmentation map $I_b$ is used in the next step of the pipeline for the detection of anomalous water bodies.

\subsection{Detection of anomalous water bodies}
Given a Sentinel-1 temporal sequence $S=\{I^1, I^2, ..., I^N\}$ containing $N$ observations, we want to compute a sequence of binary images $S_b=\{S^1_b, S^2_b, ..., S^N_b\}$ containing the anomalous water events at each image of time series.

\pparagraph{Model definition and initialization. } We propose to build a background model $B(x)$ of past observed water/non-water events for each location $x$ as in traditional change detection methods. This way, we can consider anomalous water events as \textit{changes} and exclusively segment those areas, while discarding regions permanently covered by water. It is known in the change detection literature that it is more reliable to estimate the statistical distribution of a background pixel with a small number of close values than with a large number of samples~\cite{vibe_review}. Hence, let $v(x)$ be the binary value corresponding to \textit{water} or \textit{ground} clusters at pixel $x$ of a segmentation map. Similarly to the ViBe algorithm, each pixel in the image is then modeled by a collection of $K$ previously observed \textit{water}/\textit{ground} events
\begin{equation} \label{eq:model_B}
B(x) = \{v_1, v_2, \ldots, v_K\}.
\end{equation}
The background model $B(x)$ is then initialized by applying the temporal pixel median across the $n_{init}$ initial images, and we assign the resulting value to all $K$ samples for each pixel location $x$.

\pparagraph{Flood detection and model update.} When processing images in the time series, each pixel is first classified as a flooded pixel or non-flooded pixel, and the background model $B(x)$ is then updated with the new information before moving to the next image. Thus, a new observed pixel is considered a flooded pixel only if (1) the water segmentation process has classified it as a \textit{water} event, and (2) less than $k_{min}$ water events can be found in the background model at that location. More formally, the binary classification $S^i_b(x)$ corresponding to the $i^{th}$ image from the sequence $S$ is computed as
\begin{equation} \label{eq:pixel_clasification}
S^i_b(x) = \begin{cases}
            1,  & \text{if } I^i_b(x)=0 \land \Sigma^K_{j=0}(1-B^j(x)) < k_{min} \\
            0,  & \text{elsewhere}
\end{cases},
\end{equation}
where $I^i_b(x)$ is the water segmentation result of image $I^i$ at location $x$, and $B^j(x)$ the $j^{th}$ value $v_j$ stored in $B(x)$. 

The background model is updated by introducing only the non-flooded pixels at a random position in the stack of collected observations. This is analogous in change detection to updating the background model with only the pixels classified as $background$, to avoid introducing foreground information in the background model.

\pparagraph{Implementation details and results.} We select only the VV polarization band to extract water segmentation maps, and set the parameters empirically (a different set of parameters would be derived for the VH polarization data). For the speckle filter, we use a $kernel\_size=8$, and a $threshold=0.03$ and a minimum number of components $num\_components = 20$ for the thresholding. Furthermore, we set $K=5$ samples in the background model $B$ and $k_{min}=1$. For model initialization, we use the temporal median of the first $n_{init}=30$ images. We find this ensemble of parameters to perform well across a variety of scenes. A qualitative example of the obtained results for two MMFlood test sequences can be seen in Figure~\ref{fig:qualitative_results}. As observed, the proposed method is able to detect non-permanent bodies of water throughout a sequence at a low computational cost. It is worth mentioning that $K$ determines the temporal scale considered previous to an observation and the ratio between $k_{min}$ and $K$ establishes how permissive  the model should be, i.e. how anomalous the water events should be in order to be considered a flood.

\section{Conclusion}
In this article, we illustrate the ill-posed nature of recent flood detection datasets, which can lack variability of events or require discrimination of permanent water bodies from flooded areas from a single image. Hence, we emphasize the significance of using temporal data for the flood detection problem. Furthermore, we extend the MMFlood dataset to multi-date and find that flooded areas are only partially annotated in several scenes. Lastly, we provide a simple method for flood mapping exploiting temporal information to establish a baseline method for the multi-date version of MMFlood.

\bibliographystyle{IEEEbib}
\bibliography{main}

\end{document}